%% file: paper.tex
\documentclass{juliacon}
\begin{document}

\input{header}

\maketitle

\begin{abstract}

  Over the past years, the industrial sector has seen many innovations brought about by automation. 
  Inherent in this automation is the installation of sensor networks for status monitoring and data collection. 
  One of the major challenges in these data-rich environments is how to extract and exploit information from 
  these large volume of data to detect anomalies, discover patterns to reduce downtimes and manufacturing 
  errors, reduce energy usage, predict faults/failures, effective maintenance schedules, etc. 
  To address these issues, we developed \textbf{TSML}. Its technology is based 
  on using the pipeline of lightweight filters as building blocks to process huge amount of industrial time series data in parallel.  

\end{abstract}

\section{Introduction}

\textbf{TSML}\cite{tsml2019} is a \emph{Julia}\cite{bezanson2017julia} package for time series data processing, classification, and prediction. 
It provides common API for ML (Machine Learning) libraries from Python's \emph{Scikit-Learn}, 
R's \emph{Caret}, and native \emph{Julia} MLs for seamless integration of heterogeneous 
libraries to create complex ensembles for robust time series prediction, clustering, and classification.
\textbf{TSML} has the following features:
\begin{enumerate}
\item data type clustering/classification for automatic data discovery
\item aggregation based on date/time interval
\item imputation based on symmetric Nearest Neighbors
\item statistical metrics for data quality assessment and classification input features
\item ML wrapper with more than 100+ libraries from caret, scikit-learn, and julia
\item date/value matrix conversion of 1-D time series using sliding windows to generate features for ML prediction
\item pipeline API for high-level description of the processing workflow
\item specific cleaning/normalization workflow based on data type
\item automatic selection of optimised ML model
\item automatic segmentation of time-series data into matrix form for ML training and prediction 
\item extensible architecture using just two main interfaces: fit and transform
\item meta-ensembles for automatic feature and model selection
\item support for distributed/threaded computation for scalability and speed
\end{enumerate}

The \textbf{TSML} package assumes a two-column input for any time series data composed of \emph{dates} and \emph{values}. The first part of the workflow aggregates values based on the specified date/time interval which minimizes occurrence of missing values and noise. The aggregated data is then left-joined to the complete sequence of dates in a specified date/time interval. Remaining missing values are replaced by the median/mean or user-defined aggregation function of the \textit{k}-nearest neighbors (\emph{k-NN}) where \textit{k} is the symmetric distance from the location of missing value. This approach can be called several times until there are no more missing values.

\vskip 6pt

For prediction tasks, TSML extracts the date features and 
convert the value column into matrix form parameterized by 
the size and stride of the sliding window. The final part joins
 the date features and the value matrix to serve as input to the 
 ML with the output representing the values of the time periods 
 to be predicted ahead of time.
 
 \vskip 6pt
 
\textbf{TSML} uses a pipeline which iteratively calls the \texttt{fit!} and \texttt{transform!}
families of functions relying on \emph{multiple dispatch} to dynamically select the correct algorithm from the steps outlined above. Machine learning functions in 
\textbf{TSML} are wrappers to the corresponding Scikit-Learn, Caret, and native Julia ML libraries. 
There are more than hundreds of classifiers and regression functions available using \textbf{TSML}'s common API.

\section{TSML Workflow}
\label{sec:tsmlworkflow}

All major data processing types in \textbf{TSML} are subtypes of the \texttt{Transformer}. There are two major types of transformers, namely: \emph{filters} for data processing and \emph{learners} for machine learning. Both transformers implement the \texttt{fit!} and \texttt{transform!} multi-dispatch functions. All filters are direct subtypes of the \texttt{Transformer} while all learners are subtypes of the \texttt{TSLearner}. The \texttt{TSLearner} is a direct subtype of the \texttt{Transformer}.

\vskip 6pt

Filters are normally used for pre-processing tasks such as imputation, normalization, feature extraction, feature transformation, scaling, etc.
Consequently, filters' \texttt{fit!} and \texttt{transform!} functions expect one argument which represents an input data for feature extraction or transformation. Each data type must implement \texttt{fit!} and \texttt{transform!} although in some cases, only \texttt{transform!} operation is needed. For instance, \emph{square root} or \emph{log} filters do not require any initial computation of parameters to transform their inputs. On the other hand, feature transformations such as scaling, normalization, PCA, ICA, etc. require initial computation of certain parameters in their input before applying the transformation to new datasets. In these cases, initial computations of these parameters are performed by the \texttt{fit!} function while their applications to new datasets are done by the \texttt{transform!} function. 

\vskip 6pt

Learners, on the other hand, expect two arguments (input vs output) and require training cycle to optimize their parameters for optimal \emph{input--output} mapping. The training part is handled by the \texttt{fit!} function while the prediction part is handled by the \texttt{transform!} function. 

\vskip 6pt

The \textbf{TSML} workflow borrows the idea of the  \emph{Unix} pipeline\cite{orchestra2014, combineml2016}. 
The \texttt{Pipeline} data type is also a subtype of the \texttt{Transformer} and expects two arguments: input and output. The main elements in a \textbf{TSML} pipeline are series of transformers with each performing one specific task and does it well. The series of filters are used to perform pre-processing of the input while a machine learner at the end of the pipeline is used to learn the \emph{input--output} mapping. From the perspective of using the \texttt{Pipeline} where the last component is a machine learner, the \texttt{fit!} function is the training phase while the \texttt{transform!} function is the prediction or classification phase.
\vskip 6pt

The \texttt{fit!} function in the \texttt{Pipeline} iteratively calls the \texttt{fit!} and \texttt{transform!} functions in a series of transformers. If the last transformer in the pipeline is a learner, the last transformed output from a series of filters will be used as input features for the \texttt{fit!} or training phase of the said learner.

\vskip 6pt

During the prediction task, the \texttt{transform!} function in the \texttt{Pipeline} iteratively calls the \texttt{transform!} operations in each filter and learner. The transform operation is direct application of the parameters computed during normalization, scaling, training, etc. to the new data. If the last element in the pipeline during transform is a learner, it performs prediction or classification. Otherwise, the transform operation acts as a feature extractor if they are composed of filters only.

\vskip 6pt
To illustrate, below describes the main steps in using the \textbf{TSML}.
First, we create filters for csv reading, aggregation, imputation, and data quality
assessment.

\begin{lstlisting}[language = Julia]
fname = joinpath(dirname(pathof(TSML)),
  "../data/testdata.csv")
csvfilter = CSVDateValReader(Dict(
  :filename=>fname,
  :dateformat=>"dd/mm/yyyy HH:MM"))
valgator = DateValgator(Dict(
  :dateinterval=>Dates.Hour(1)))
valnner = DateValNNer(Dict(
  :dateinterval=>Dates.Hour(1)))
stfier = Statifier(Dict(:processmissing=>true))
\end{lstlisting}

\begin{figure}[htbp]
   \centering
   \includegraphics[width=\columnwidth]{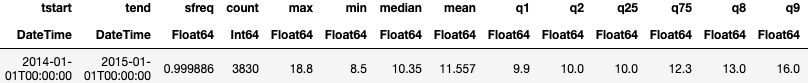} 
   \vskip 2pt
      \includegraphics[width=\columnwidth]{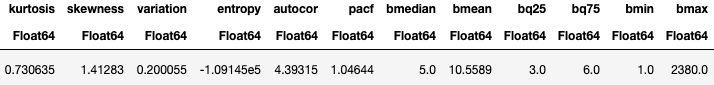} 
   \caption{Data Quality Statistics. Column names starting with `b` refer to statistics of contiguous blocks of missing data.}
   \label{fig:dataquality}
\end{figure}

We can then setup a pipeline containing these filters to process the csv data
by aggregating the time series hourly and check the data quality using the
\texttt{Statifier} filter (Fig.~\ref{fig:dataquality}).

\begin{lstlisting}[language = Julia]
apipeline = Pipeline(Dict(
  :transformers => [csvfilter, valgator, stfier]))
fit!(apipeline)
mystats = transform!(apipeline)
@show mystats 
\end{lstlisting}

A mentioned previously, the \texttt{fit!} and \texttt{transform!} in the pipeline
iteratively calls the corresponding \texttt{fit!} and \texttt{transform!} within each filter. 
This common API relying on \emph{Julia's} multi-dispatch mechanism greatly simplifies the implementations, operations, 
and understanding of the entire workflow. In addition, extending \textbf{TSML} functionality is just a 
matter of creating a new data type filter and define its own  \texttt{fit!} and \texttt{transform!} 
functions.

\vskip 6pt

\begin{figure}[htbp]
   \centering
   \includegraphics[width=\columnwidth]{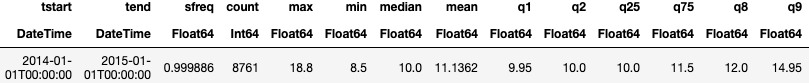} 
   \vskip 2pt
      \includegraphics[width=\columnwidth]{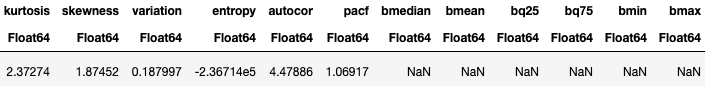} 
   \caption{Statistics after Imputation. All statistics for blocks of missing data indicate NaN to indicate the stats of empty set. This implies that all missing blocks are imputed.}
   \label{fig:imputation}
\end{figure}

In the \texttt{Statifier} filter result, blocks of missing data is indicated by column names starting
with \emph{b}. Running the code indicates that there are plenty of missing data blocks.
We can add the \texttt{ValNNer} filter to perform \emph{k}-nearest neighbour (\emph{k-NN}) imputation and check
the statistics (Fig.~\ref{fig:imputation}):

\begin{lstlisting}[language = Julia]
bpipeline = Pipeline(Dict(
  :transformers => [csvfilter, valgator, 
                    valnner,stfier]))
fit!(bpipeline)
imputed = transform!(bpipeline)
@show imputed
\end{lstlisting}

The result in Fig.~\ref{fig:imputation} indicates \emph{NaN} for all missing data statistics column because the set 
of missing blocks count is now empty.

\vskip 6pt

\begin{figure}[htbp]
   \centering
   \includegraphics[width=\columnwidth]{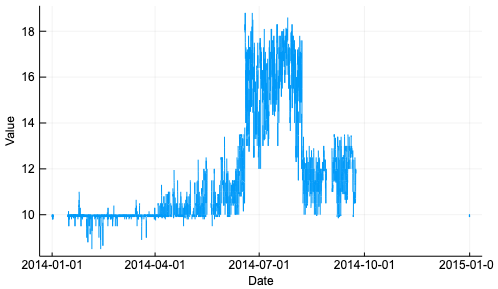} 
   \caption{Plot with Missing Data}
   \label{fig:mplot}
\end{figure}

We can also visualise our time series data using the \texttt{Plotter} filter instead of the \texttt{Statifier} as shown in Fig~\ref{fig:mplot}. Looking closely, you will see discontinuities in the plot due to blocks of missing data.

\begin{lstlisting}[language = Julia]
pltr=Plotter(Dict(:interactive => true))
plpipeline = Pipeline(Dict(
  :transformers => [csvfilter, valgator, pltr]))
fit!(plpipeline)
transform!(plpipeline)
\end{lstlisting}

\begin{figure}[htbp]
   \centering
   \includegraphics[width=\columnwidth]{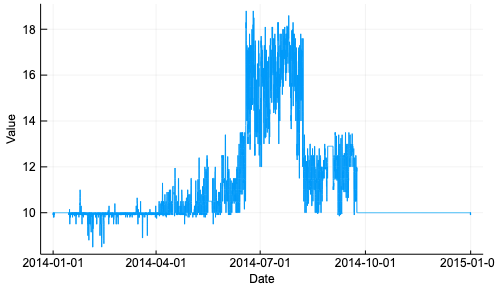} 
   \caption{Plot after Data Imputation}
   \label{fig:amplot}
\end{figure}

Using the imputation pipeline described before, we can visualise the result by replacing the \texttt{Statifier} with the \texttt{Plotter} filter. Figure~\ref{fig:amplot} shows the plot after imputation which gets rid of missing data.

\begin{lstlisting}[language = Julia]
bplpipeline = Pipeline(Dict(
  :transformers => [csvfilter, valgator, 
                    valnner,pltr]))
fit!(bplpipeline)
transform!(bplpipeline)
\end{lstlisting}

\subsection{Processing Monotonic Time Series}
This subsection dicusses additional filters to handle monotonic data which are commonly employed in energy/water meter and footfall sensors. In the former case, the time series type is strictly monotonically increasing while in the latter case, the monotonicity happens daily. We use the filter called \texttt{Monotonicer} which automatically detects these two types of monotonic sensors and apply the normalisation accordingly. 

\begin{lstlisting}[language = Julia]
mono = joinpath(dirname(pathof(TSML)),
  "../data/typedetection/monotonic.csv")
monocsv = CSVDateValReader(Dict(:filename=>mono,
  :dateformat=>"dd/mm/yyyy HH:MM"))
\end{lstlisting}

\begin{figure}[htbp]
   \centering
   \includegraphics[width=0.9\columnwidth]{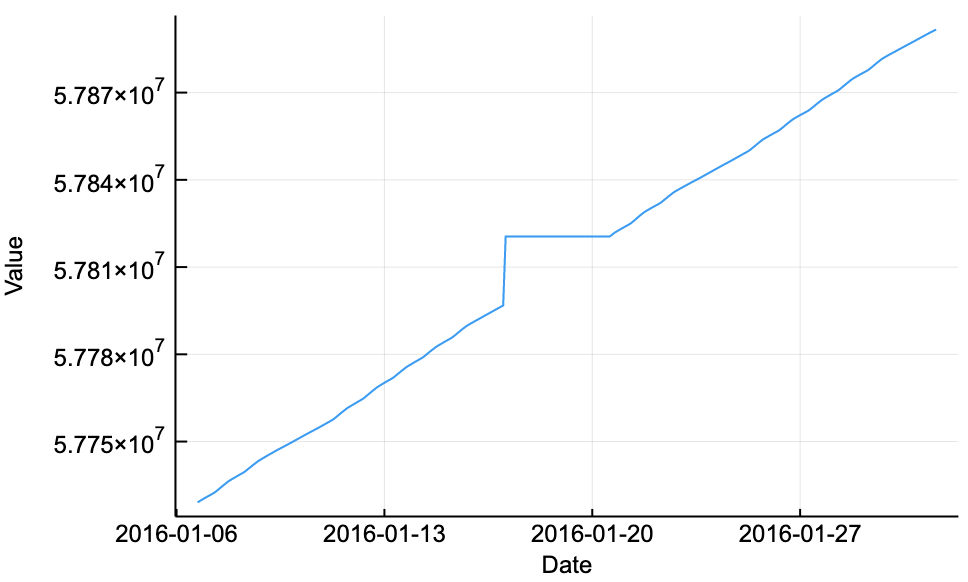} 
   \caption{Monotonic Time Series}
   \label{fig:mono}
\end{figure}

Let us plot in Fig.~\ref{fig:mono} the monotonic data with the usual workflow of aggregating and imputing the data first.

\begin{lstlisting}[language = Julia]
monopipeline = Pipeline(Dict(
  :transformers => [monofilecsv,valgator,
                    valnner,pltr]))
fit!(monopipeline)
transform!(monopipeline)
\end{lstlisting}

\begin{figure}[htbp]
   \centering
   \includegraphics[width=\columnwidth]{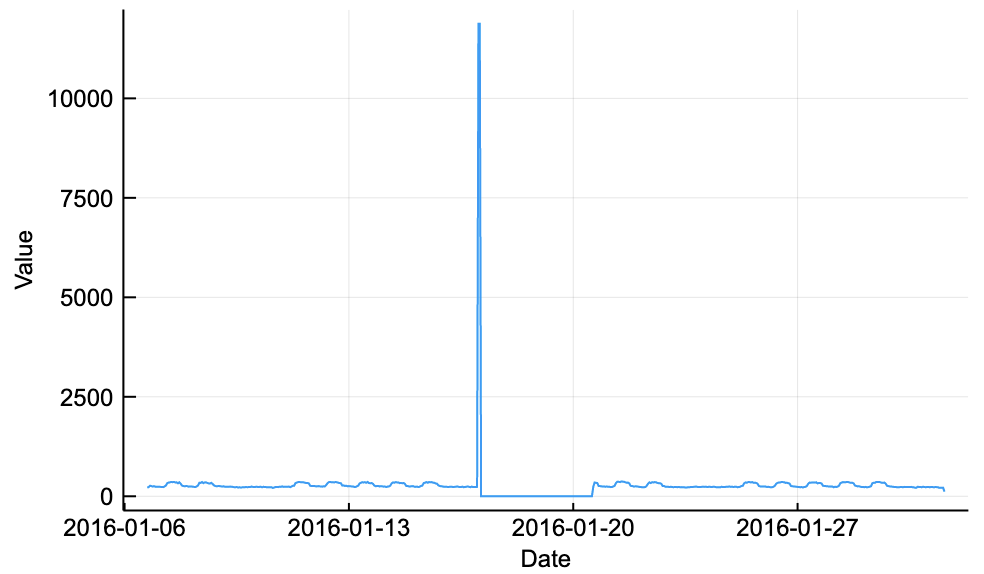} 
   \caption{Normalized Monotonic Time Series}
   \label{fig:nmono}
\end{figure}

Let us now normalise using \texttt{Monotonicer} and show the plot in Fig.~\ref{fig:nmono}.

\begin{lstlisting}[language = Julia]
mononicer = Monotonicer(Dict())
monopipeline = Pipeline(Dict(
  :transformers => [monofilecsv,valgator,valnner,
                    mononicer, pltr]))
fit!(monopipeline) 
transform!(monopipeline)
\end{lstlisting}

\begin{figure}[htbp]
   \centering
   \includegraphics[width=\columnwidth]{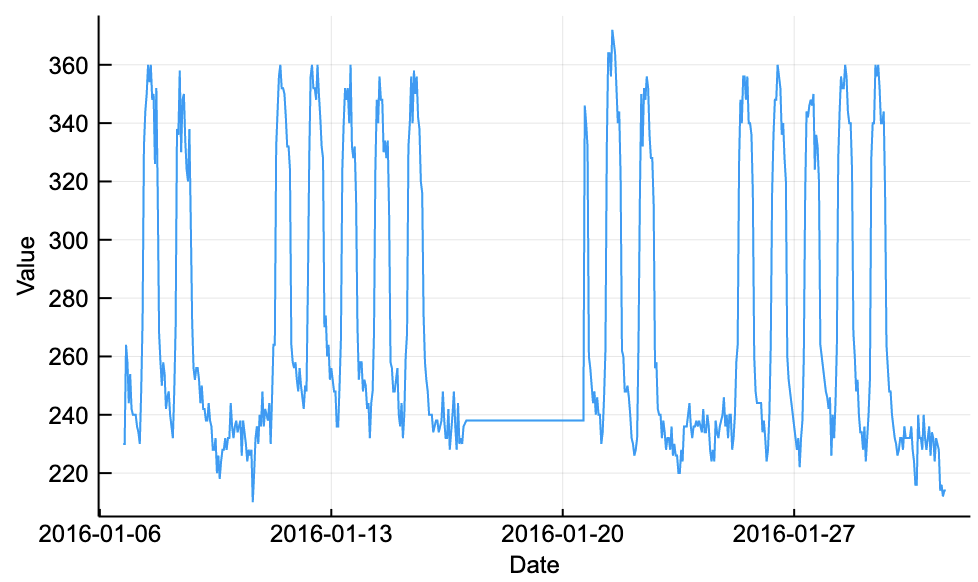} 
   \caption{Normalized Monotonic Time Series With Outlier Removal Filter}
   \label{fig:outnicer}
\end{figure}

The presence of outlier due to some random errors during meter reading  becomes obvious after the normalisation. To remedy this issue, we add the \texttt{Outliernicer} filter which detects outliers and replace them  using the \emph{k-NN} imputation technique used by the \texttt{DateValNNer} filter (Fig.~\ref{fig:outnicer}).

\begin{lstlisting}[language = Julia]
outliernicer = Outliernicer(
       Dict(:dateinterval=>Dates.Hour(1)))
monopipeline = Pipeline(Dict(
  :transformers => [monofilecsv,valgator,valnner,
                    mononicer,outliernicer,pltr]))
fit!(monopipeline)
transform!(monopipeline)
\end{lstlisting}

\subsection{Processing Daily Monotonic Time Series}
\begin{figure}[htbp]
   \centering
   \includegraphics[width=\columnwidth]{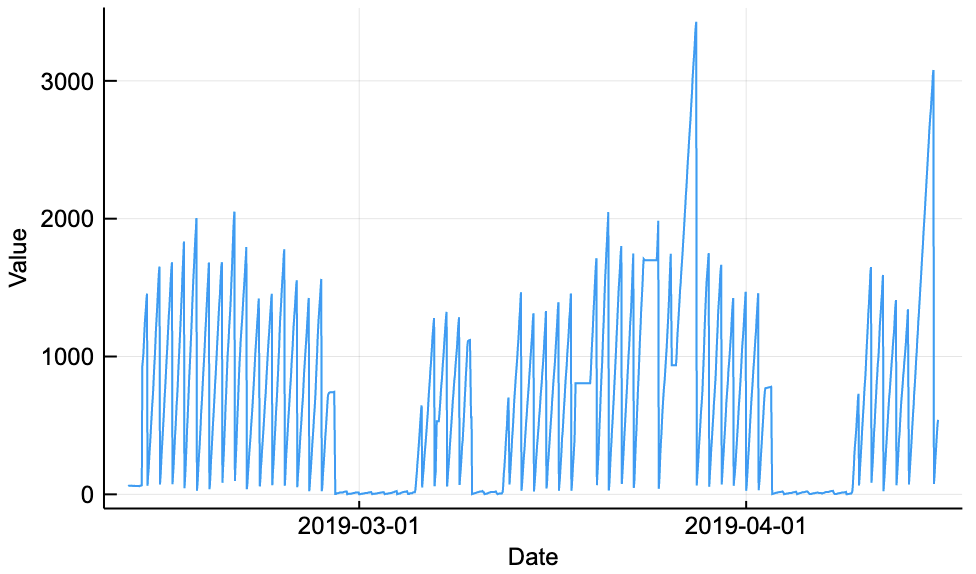} 
   \caption{Daily Monotonic Time Series}
   \label{fig:dailymono}
\end{figure}

We follow similar workflow in the previous subsection to normalize daily monotonic time series. First, let us visualize the original data after aggregation and imputation (Fig.~\ref{fig:dailymono}). 

\begin{lstlisting}[language = Julia]
dailymono = joinpath(dirname(pathof(TSML)),
     "../data/type-detection/dailymonotonic.csv")
dailymonocsv = CSVDateValReader(Dict(
     :filename=>dailymono,
     :dateformat=>"dd/mm/yyyy HH:MM"))
dailymonopipeline = Pipeline(Dict(
  :transformers => [dailymonocsv,valgator,
                    valnner,pltr]))
fit!(dailymonopipeline)
transform!(dailymonopipeline)
\end{lstlisting}

\vskip 6pt

\begin{figure}[htbp]
   \centering
   \includegraphics[width=\columnwidth]{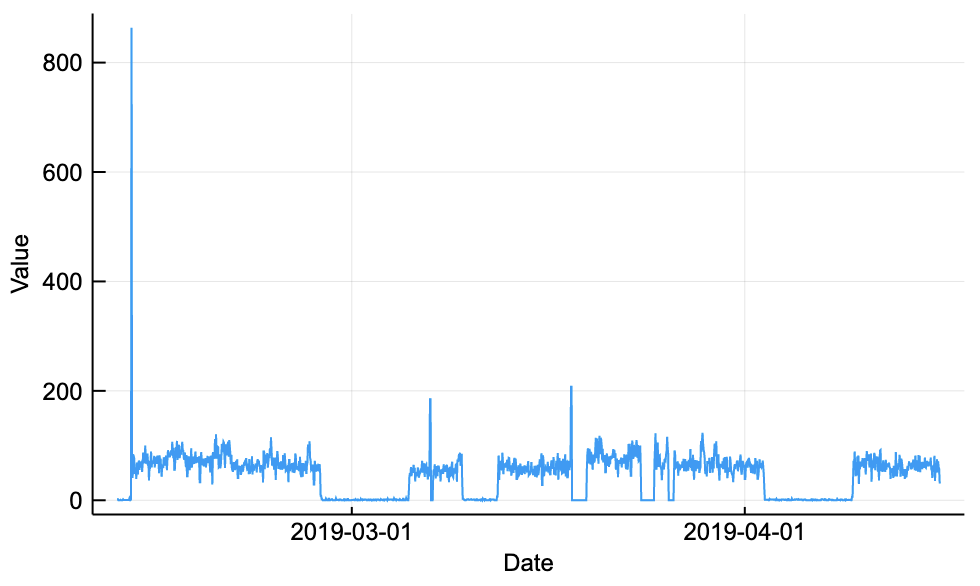}  
   \caption{Normalized Daily Monotonic Time Series}
   \label{fig:ndailymono}
\end{figure}

Then, we reuse the \texttt{Monotonicer} filter in the previous subsection to normalise the data and plot (Fig.~\ref{fig:ndailymono}).

\begin{lstlisting}[language = Julia]
dailymonopipeline = Pipeline(Dict(
  :transformers => [dailymonocsv,valgator,
                    valnner,mononicer,pltr]))
fit!(dailymonopipeline)
transform!(dailymonopipeline)
\end{lstlisting}

\begin{figure}[htbp]
   \centering
   \includegraphics[width=\columnwidth]{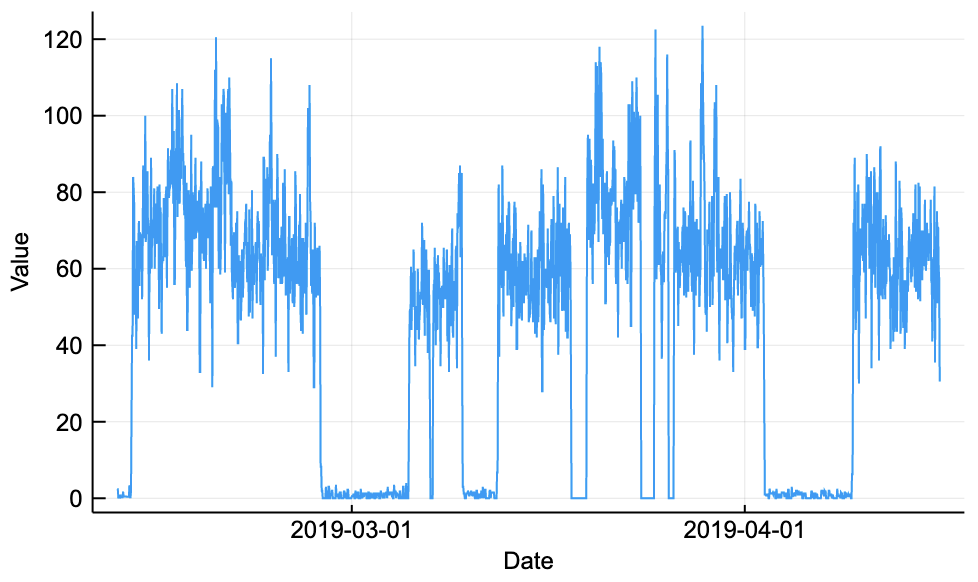}  
   \caption{Normalized Daily Monotonic Time Series with Outlier Detector}
   \label{fig:outndailymono}
\end{figure}

To remove outliers, we can reuse the \texttt{Outliernicer} filter in the previous subsection and plot the cleaned data (Fig.~\ref{fig:outndailymono}).

\begin{lstlisting}[language = Julia]
dailymonopipeline = Pipeline(Dict(
  :transformers=>[dailymonocsv,valgator,valnner,
                  mononicer,outliernicer,pltr]))
fit!(dailymonopipeline)
transform!(dailymonopipeline)
\end{lstlisting}

\section{Time Series Classification}

We can use the knowledge we learned in setting up the \textbf{TSML}
\emph{pipeline} containing \emph{filters} and \emph{machine learners} to build higher level
operations to solve a specific industrial problem. One major problem
which we consider relevant because it is a common issue in IOT (Internet of Things) 
 is the time series classification. This problem is prevalent nowadays 
due to the increasing need to use many sensors to monitor status in different aspects of industrial
operations and maintenance of cars, buildings, hospitals, supermarkets, homes, and cities.

\vskip 6pt

Rapid deployment of these sensors result to many of them not properly labeled or classified.
Time series classification is a significant first step for optimal prediction and anomaly detection.
Identifying the correct sensor data types can help in the choice of what the most optimal prediction 
model to use for actuation or pre-emption to minimise wastage in resource utilisation.
To successfully perform the latter operations, it is necessary to identify first the time series
type so that appropriate model and cleaning routines can be selected for optimal model performance. The  \texttt{TSClassifier} filter aims to address this problem and its usage is described below.

\vskip 6pt

First, we setup the locations of files for training, testing, and saving the model.
Next, we start the training phase by calling \texttt{fit!} which loads
file in the training directory and learn the mapping between their
statistic features extracted by \texttt{Statifier} with their types indicated
by a substring in their filenames. Once the training is done, the final model
is saved in the \emph{model} directory which will be used for 
testing accuracy and classifying new time series datasets. 

\vskip 6pt

The code below initialises the \texttt{TSClassifier} with the locations of the \emph{training}, \emph{testing}, and \emph{model} repository. Training is carried out by the \texttt{fit!} function which extracts the stat features of the training data and save them as a \emph{dataframe}  to be processed by the \emph{RandomForest} classifier. The trained model is saved in the \emph{model} directory and used during testing.

\begin{lstlisting}[language = Julia]
trdirname = joinpath(dirname(pathof(TSML)),
     "../data/realdatatsclassification/training")
tstdirname = joinpath(dirname(pathof(TSML)),
     "../data/realdatatsclassification/testing")
modeldirname = joinpath(dirname(pathof(TSML)),
     "../data/realdatatsclassification/model")
tscl = TSClassifier(Dict(
  :trdirectory=>trdirname,
  :tstdirectory=>tstdirname,
  :modeldirectory=>modeldirname,
  :num_trees=>75)
)
fit!(tscl)
predictions = transform!(tscl)
@show testingAccuracy(predictions)
\end{lstlisting}

\begin{figure}[htbp]
   \centering
   \includegraphics[width=0.5\columnwidth]{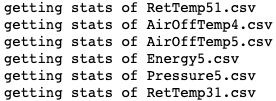} 
   
   ..........
   
   ..........
   \vskip 2pt
   \includegraphics[width=0.4\columnwidth]{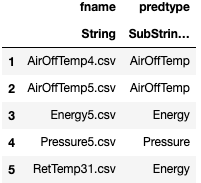} 
   \caption{Extraction of Statistical Features for Training and Prediction. By convention, the \texttt{TSClassifier} validates the ground truth based on the filename which contains the label of the sensor type disregarding the numerical component.}
   \label{fig:tcl}
\end{figure}

Figure~\ref{fig:tcl} shows: a) a snapshot of the output during training and testing which extracts the statistical features of the time series; and b) the testing performance of the classifier. The training and testing data are labeled based on their sensor type for easier validation. The labels are not used as input during training. The classification workflow is purely driven by the statistical features. The prediction indicates 80\% accuracy.

\section{Extending TSML with Scikit-Learn and Caret}
In the latest \textbf{TSML} version (2.3.4 and above), 
we refactored the base \textbf{TSML}
to only include pure \emph{Julia} code implementations and moved the external libs and binary dependencies into the \textbf{TSMLextra} package. 
One major reason is to have a smaller code base so that it can be easily
maintained and rapidly deployed in a dockerized solution for \emph{Kubernetes} or IBM's \emph{OpenShift} cluster. Moreover, smaller codes make
static compilation fast for smaller docker image  
in cloud deployment. 

\vskip 6pt

There are cases where the main task of time series classification 
requires more complex ensemble model using hierarchy or tree structure where 
members are composed of heterogeneous ML learners derived from binaries in 
different languages. For illustration purposes, we will show how to 
ensemble ML libraries from \emph{Scikit-Learn} and \emph{Caret} using \textbf{TSML} meta-ensembles that support the  \texttt{fit!} and \texttt{transform!} APIs.

\subsection{Parallel TSML Using Distributed Workflow}

We will use \emph{Julia's} built-in support for parallelism by using the \emph{Distributed} standard library. We also let \emph{Julia} detect the number of processors available and activate them using the following statements:

\begin{lstlisting}[language = Julia]
using Distributed 
nprocs() == 1 && addprocs()
\end{lstlisting}

With several workers active, we use the \emph{@everywhere} macro to load the necessary filters and transformers to all workers.

\begin{lstlisting}[language = Julia]
@everywhere using TSML
@everywhere using TSMLextra
@everywhere using DataFrames
@everywhere using Random
@everywhere using Statistics
@everywhere using StatsBase: iqr
@everywhere using RDatasets
\end{lstlisting}

With all the necessary \textbf{TSML} functions loaded, we can now setup the different MLs starting with some learners from \emph{Caret} and \emph{Scikit-Learn}. The list is not exhaustive for demonstration purposes.

\begin{lstlisting}[language = Julia]
# Caret ML
@everywhere caret_svmlinear = 
   CaretLearner(Dict(:learner=>"svmLinear"))
@everywhere caret_treebag = 
   CaretLearner(Dict(:learner=>"treebag"))

# Scikit-Learn ML
@everywhere sk_knn = 
  SKLearner(Dict(:learner=>"KNeighborsClassifier"))
@everywhere sk_gb = 
  SKLearner(Dict(:learner=>
    "GradientBoostingClassifier",
    :impl_args=>Dict(:n_estimators=>10)))
@everywhere sk_extratree = 
  SKLearner(Dict(:learner=>"ExtraTreesClassifier",
    :impl_args=>Dict(:n_estimators=>10)))
@everywhere sk_rf = 
  SKLearner(Dict(:learner=>
    "RandomForestClassifier",
    :impl_args=>Dict(:n_estimators=>10)))
\end{lstlisting}

Let us setup ML instances from a pure \emph{Julia} implementation of learners and ensembles wrapped from the \emph{DecisionTree.jl} package \cite{decisiontree2008,orchestra2014,combineml2016,tsmlextra2019}.

\begin{lstlisting}[language = Julia]
# Julia ML
@everywhere jrf = RandomForest()
@everywhere jpt = PrunedTree()
@everywhere jada = Adaboost()

# Julia Ensembles
@everywhere jvote_ens=VoteEnsemble(Dict(
   :learners=>[jrf,jpt,sk_gb,sk_extratree,sk_rf]))
@everywhere jstack_ens=StackEnsemble(Dict(
   :learners=>[jrf,jpt,sk_gb,sk_extratree,sk_rf]))
@everywhere jbest_ens=BestLearner(Dict(
   :learners=>[jrf,sk_gb,sk_rf]))
@everywhere jsuper_ens=VoteEnsemble(Dict(
   :learners=>[jvote_ens,jstack_ens,
               jbest_ens,sk_rf,sk_gb]))
\end{lstlisting}

Next, we setup the pipeline for training and prediction.

\begin{lstlisting}[language = Julia]
@everywhere function predict(learner,
            data,train_ind,test_ind)        
  features = convert(Matrix,data[:, 1:(end-1)])
  labels = convert(Array,data[:, end])
  # Create pipeline
  pipeline = Pipeline(
    Dict(
      :transformers => [
        OneHotEncoder(), # nominal to bits
        Imputer(), # Imputes NA values
        StandardScaler(), # normalize
        learner # Predicts labels on instances
      ]
    )
  )
  # Train
  fit!(pipeline, features[train_ind, :],
       labels[train_ind])  
  # Predict
  predictions = transform!(pipeline, 
      features[test_ind, :])
  # Assess predictions
  result = score(:accuracy, 
      labels[test_ind], predictions)
  return result
end
\end{lstlisting}

Finally, we setup the \texttt{parallelmodel} function to run different learners distributed to different workers running in parallel relying on \emph{Julia's} native support of parallelism. Take note that there are two parallelisms in the code. The first one is the distribution of task in different trials and the second one is the distribution of tasks among different models for each trial. It is interesting to note that with this relative compact function definition, the \emph{Julia} language makes it easy to define a parallel task within another parallel task 
in a straightforward manner without any problem.

\begin{lstlisting}[language = Julia]
function parallelmodel(learners::Dict,
         data::DataFrame;trials=5)
  models=collect(keys(learners))
  ctable=@distributed (vcat) for i=1:trials
    # Split into training and test sets
    Random.seed!(3i)
    (trndx, tstndx) = holdout(size(data, 1), 0.20)
    acc=@distributed (vcat) for model in models
      res=predict(learners[model],
             data,trndx,tstndx)
      println("trial ",i,", ",model," => ",
              round(res))
      [model res i]
    end
    acc
  end
  df = ctable |> DataFrame
  rename!(df,:x1=>:model,:x2=>:acc,:x3=>:trial)
  gp=by(df,:model) do x
    DataFrame(mean=mean(x.acc),std=std(x.acc),
              n=length(x.acc)) 
  end
  sort!(gp,:mean,rev=true)
  return gp
end
\end{lstlisting}

We benchmark the performance of the different machine learners by creating a dictionary of workers containing instances of learners from \emph{Caret}, \emph{Scikit-Learn}, and \emph{Julia} libraries. We pass the dictionary of learners to the \texttt{parallelmodel} function for evaluation.

\begin{lstlisting}[language = Julia]
learners=Dict(
  :jvote_ens=>jvote_ens,:jstack_ens=>jstack_ens,
  :jbest_ens=>jbest_ens,:jrf=>jrf,:jada=>jada,
  :jsuper_ens=>jsuper_ens, 
  :crt_svmlinear=>caret_svmlinear,
  :crt_treebag=>caret_treebag,
  :skl_knn=>sk_knn,:skl_gb=>sk_gb,
  :skl_extratree=>sk_extratree, :sk_rf=>sk_rf
)

datadir = joinpath("tsdata/")
tsdata = extract_features_from_timeseries(datadir)
first(tsdata,5)

respar = parallelmodel(learners,tsdata;trials=3)
\end{lstlisting}

\begin{figure}[htbp]
   \centering
   \includegraphics[width=0.8\columnwidth]{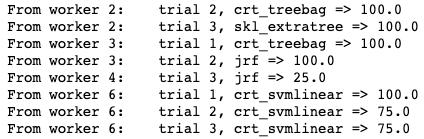} 
   \caption{Output During Training with Several Workers}
   \label{fig:sim}
\end{figure}

\begin{figure}[htbp]
   \centering
   \includegraphics[width=0.6\columnwidth]{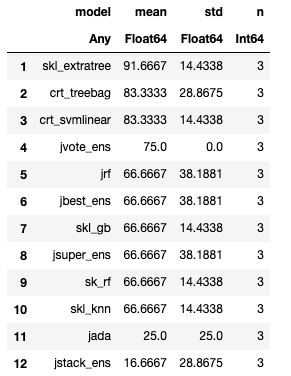} 
   \caption{@distributed: Classification Performance in 3 Trials}
   \label{fig:performance}
\end{figure}

The data used in the experiment are sample snapshots of the data in our building operations. For reproducibility, the data can be found in the  \emph{juliacon2019-paper} branch of \textbf{TSML} in \emph{Github}:  /data/benchmark/tsclassifier. There are four time series types, namely: AirOffTemp, Energy, Pressure, and RetTemp.  We  took a minimal number of samples and classes for the sake of discussion and demonstration purposes in this paper.

\vskip 6pt

Figures~\ref{fig:sim}~and~\ref{fig:performance}  show a snapshot of running workers exploiting the distributed library of \emph{Julia} and the classification performance of each model, respectively. There are 8 workers running in parallel over 12 different machine learning classifiers. 

\vskip 6pt

From the results, \emph{ExtraTree} from \emph{Scikit-Learn} has the best performance with 91.67\% accuracy followed by \emph{TreeBag} and \emph{SVMLinear} from \emph{Caret} library with 83.33 \% accuracy for both. With this workflow, it becomes trivial to search for optimal model by running them in parallel relying on \emph{Julia} to do the low-level tasks of scheduling and queueing as well as making sure that  the dynamically available compute resources such as cpu cores and memory resources are fairly optimised.           

\subsection{Parallel TSML Using Threads Workflow}
With \emph{Julia} 1.3, lightweight multi-threading support in \emph{Julia} becomes possible. We will be using the pure \emph{Julia}-written ML models because installing external dependencies such as \emph{Caret} MLs through \emph{RCall} package has some issues with the alpha version of \emph{Julia} 1.3 at this point in time. We will update this documentation and add more MLs once the issues are resolved.

\vskip 6pt

The main difference in the workflow between \emph{Julia's} distributed computation model compared to the threaded model is the presence of \texttt{@everywhere} macro in the former for each function defined to indicate that these function definitions shall be exported to all running workers. Since threaded processes share the same memory model with the \emph{Julia} main process, there is no need for this macro. Instead, threading workflow requires the use of \emph{ReentrantLock} in the update of the global dataframe that accumulates the prediction performance of models running in their respective threads. In similar observation with the distributed framework, the \texttt{threadedmodel} function contains two parallelism: threads in different trials and threads among models in each trial. The function is surprisingly compact to implement threads within threads without issues and the main bottleneck happens only during the update operation of the global \emph{ctable} dataframe.

\begin{lstlisting}[language = Julia]
function threadedmodel(learners::Dict,
         data::DataFrame;trials=5)
  Random.seed!(3)
  models=collect(keys(learners))
  global ctable = DataFrame()
  @threads for i=1:trials
     # Split into training and test sets
     (train_ind, test_ind) = 
          holdout(size(data, 1), 0.20)
     mtx = SpinLock()
     @threads for themodel in models
       res=predict(learners[themodel],
          data,train_ind,test_ind)
       println(themodel," => ",round(res),", 
          thread=",threadid())
       lock(mtx)
       global ctable=vcat(ctable,
          DataFrame(model=themodel, acc=res))
       unlock(mtx)
     end
  end
    df = ctable |> DataFrame
    gp=by(df,:model) do x
       DataFrame(mean=mean(x.acc),
          std=std(x.acc),n=nrow(x))
    end
    sort!(gp,:mean,rev=true)
    return gp
end
\end{lstlisting}

Let us define a set of learners that are written in pure \emph{Julia} for this thread experiment.

\begin{lstlisting}[language = Julia]
# Julia ML
jrf = RandomForest(Dict(:impl_args=>
   Dict(:num_trees=>500)))
jpt = PrunedTree()
jada = Adaboost(Dict(:impl_args=>
   Dict(:num_iterations=>20)))

## Julia Ensembles
jvote_ens=VoteEnsemble(Dict(:learners=>
   [jrf,jpt,jada]))
jstack_ens=StackEnsemble(Dict(:learners=>
   [jrf,jpt,jada]))
jbest_ens=BestLearner(Dict(:learners=>
   [jrf,jpt,jada]))
jsuper_ens=VoteEnsemble(Dict(:learners=>
   [jvote_ens,jstack_ens,jbest_ens]));
\end{lstlisting}

Let us run in parallel the different models using the same dataset with that of the distributed workflow.

\begin{lstlisting}[language = Julia]
using Base.Threads

learners=Dict(
      :jvote_ens=>jvote_ens,
      :jstack_ens=>jstack_ens,
      :jbest_ens=>jbest_ens,
      :jrf => jrf,:jada=>jada,
      :jsuper_ens=>jsuper_ens);
      
datadir = joinpath("tsdata/")
tsdata = extract_features_from_timeseries(datadir)

resthr = threadedmodel(learners,tsdata;trials=10)
\end{lstlisting}

\begin{figure}[htbp]
   \centering
   \includegraphics[width=0.5\columnwidth]{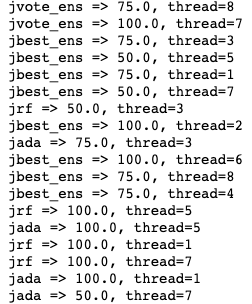} 
   \caption{Output During Training Using Several Threads}
   \label{fig:threadrunning}
\end{figure}

\begin{figure}[htbp]
   \centering
   \includegraphics[width=0.5\columnwidth]{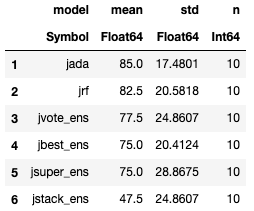} 
   \caption{@threads: Classification Performance in 10 Trials}
   \label{fig:threadresult}
\end{figure}

Figures~\ref{fig:threadrunning}~and~\ref{fig:threadresult} show example snapshot of the running threads and the final result of classification, respectively. In this experiment, Adaboost has 85.0\% accuracy followed by Random Forest with 82.5\% accuracy.

\section{Applications Using Public IoT Datasets}

This section summarizes the results of applying the TSML workflow for classification tasks using data in this site: 
\emph{http://www.timeseriesclassification.com/}. Among the hundreds of datasets available, we handpicked 
only 4 datasets, namely: ElectricDevices, RefrigerationDevices, FordB, and Earthquakes. 

\vskip 6pt

ElectricDevices dataset covers 251 households, sampled in 2-minute intervals over a month. The target is to collect 
how consumers use electricity within the home based on the devices they use at a particular time of the day.  The dataset
is highly imbalanced with the following total count in each class, repectively: 1394, 4187, 1606, 2639, 4275, 1252, 1284. This dataset is a good test on the robustness of the classifier to handle bias during training. The class distribution in training is consist of: 727, 2231, 851, 1471, 2406, 509, 728. For testing, the class distribution is: 667, 1956,  755, 1165, 1869,  743,  556. Using this highly imbalanced dataset is a good way to determine which among the classifiers has the best method to deal with this sampling bias. There are 96 input features and 6 output classes in a total of 16637  samples. 

\vskip 6pt

RefrigerationDevices dataset is based on similar study with that of ElectricalDevices dataset. The dataset uses the same set of households and sampling protocol for gathering electric consumption data from three types of refrigeration devices: Fridge/Freezer, Refrigerator, Upright Freezer. Unlike in ElectricDevices, the class distribution in RefrigerationDevices are balanced, i.e., 250 for each class in training and 125 for each class during testing. There are 720 input features and 3 output classes in a total of 
750 samples.

\vskip 6pt

The Earthquakes classification problem requires predicting whether a major event is about to occur based on the most recent readings in the surrounding area. The data is aggregated hourly with the Rictor scale reading over 5 indicating a  major event. There are 368 negative cases versus 93 positive cases of major events. This dataset is highly imbalanced in class distribution which makes predicting whether a major event occurs very problematic. The training set is composed of 264 negative examples and only 58 positive examples. The testing data is composed of 104 negative cases and only 35 positive cases.
There are 512 input features and 2 output classes in a total of 461 samples. The highly imbalanced distribution of this dataset will provide a good way to determine which among the classifiers are robust to deal with this bias.

\vskip 6pt

FordB dataset is a classification problem to diagnose whether a certain symptom exists in an automative system. There are 500 measurements of engine noise. The training data were collected in typical operating conditions while the test data were collected under noisy conditions. The dataset is slightly imbalanced consisting of 1860 class1 vs 1776 class2 in training while 401 class1 and 409 class2 in testing. There are 500 input features and 2 output classes in a total of 4446 samples.

\vskip 6pt

\begin{table}
\tbl{Classification Problems}{
\begin{tabular}{|r|c|c|c|}\hline
\textbf{Problem} & \textbf{Classes} & \textbf{Imbalance} & \textbf{Dimension}\\\hline
Electric Devices & 6 & Y & 16637 x 96\\\hline
Refrigeration Devices & 3 & N & 750 x 720\\\hline
Earthquakes & 2 & Y & 461 x 512\\\hline
FordB & 2 & N & 4446 x 500 \\\hline
\end{tabular}}
\label{tab:balance}
\end{table}

Table~\ref{tab:balance} summarizes the differences and similarities among the different classification problems under consideration. Two of the problems have highly imbalanced dataset, namely: Electric Devices and Earthquakes. Both FordB and Refrigeration Devices have no or slightly less imbalanced data. 

\vskip 6pt

One thing to note is that the Refrigeration dataset has almost the same number of features with its total number of samples. We expect that this will result into a much harder classification problem eventhough it does not suffer from imbalanced class distribution. Ideally, there must be more samples than features in order for the classifier to properly extract the correct subset of features for robust classification.

\subsection{Results}

\begin{table}
\tbl{Electric Devices}{
\begin{tabular}{|r|c|c|c|}\hline
\textbf{Model} & \textbf{MeanFscore} & \textbf{Std} & \textbf{Trials}\\\hline
c-rf & 0.56 & 0.00 & 10\\\hline
c-treebag & 0.54 & 0.01 & 10\\\hline
s-extratree & 0.52 & 0.01 & 10\\\hline
j-vote & 0.51 & 0.01 & 10 \\\hline
j-rf & 0.51 & 0.01 & 10 \\\hline
... & ... & ... & ...\\\hline
\end{tabular}}
\label{tab:elec}
\end{table}

\begin{table}
\tbl{Refrigeration Devices}{
\begin{tabular}{|r|c|c|c|}\hline
\textbf{Model} & \textbf{MeanFscore} & \textbf{Std} & \textbf{Trials}\\\hline
s-gb & 0.51 & 0.00 & 10\\\hline
c-rf & 0.51 & 0.01 & 10\\\hline
c-treebag & 0.50 & 0.01 & 10\\\hline
j-super & 0.50 & 0.02 & 10 \\\hline
s-rf & 0.48 & 0.03 & 10 \\\hline
... & ... & ... & ...\\\hline
\end{tabular}}
\label{tab:ref}
\end{table}

\begin{table}
\tbl{Earthquakes}{
\begin{tabular}{|r|c|c|c|}\hline
\textbf{Model} & \textbf{MeanFscore} & \textbf{Std} & \textbf{Trials}\\\hline
c-treebag & 0.86 & 0.01 & 10\\\hline
c-rf & 0.86 & 0.00 & 10\\\hline
c-rpart & 0.86 & 0.00 & 10\\\hline
s-extratree & 0.85 & 0.01 & 10 \\\hline
s-gb & 0.85 & 0.00 & 10 \\\hline
... & ... & ... & ...\\\hline
\end{tabular}}
\label{tab:earth}
\end{table}

\begin{table}
\tbl{FordB}{
\begin{tabular}{|r|c|c|c|}\hline
\textbf{Model} & \textbf{MeanFscore} & \textbf{Std} & \textbf{Trials}\\\hline
c-rf & 0.66 & 0.01 & 10\\\hline
s-knn & 0.62 & 0.00 & 10\\\hline
j-vote & 0.61 & 0.01 & 10\\\hline
j-rf & 0.61 & 0.01 & 10 \\\hline
j-super & 0.60 & 0.03 & 10 \\\hline
... & ... & ... & ...\\\hline
\end{tabular}}
\label{tab:fordb}
\end{table}

Due to the data imbalance, the typical accuracy measurement will not be able to capture the performance of the algorithms because its value may be overshadowed by the dominant class. In this regard, we use F-score to measure the performance of a given classifier to each of its classes and get the mean of these F-scores.

\vskip 6pt

Table~\ref{tab:elec} shows the top 5 performing classifiers together with the 2 worst performing classifiers. The first letter of each classifier's name indicates whether the classifier comes from (c)aret, (s)cikitLearn, or (j)ulia. The table indicates that \emph{Random Forest} and \emph{TreeBag} from \emph{Caret} performed the best followed by \emph{ExtraTree} of \emph{ScikitLearn}. Julia's \emph{Vote} ensemble and \emph{Random Forest} complete the top 5. On the other hand, Julia's \emph{Adaboost} and Caret's \emph{RPart} are the worst classifiers for this problem.  

\vskip 6pt

Using similar naming convention, Table~\ref{tab:ref} indicates that \emph{GradientBoost} from \emph{ScikitLearn} and \emph{Random Forest} from \emph{Caret} are the best classifiers for the Refrigeration Devices problem. The two worst algorithms are the same as in Table~\ref{tab:elec}.

\vskip 6pt

For Earthquakes classification problem (Table~\ref{tab:earth}), the top 3 best classifiers are dominated by those from \emph{Caret} library, namely: \emph{TreeBag}, \emph{Random Forest}, and \emph{RPart}. The worst performers are from \emph{Caret's} \emph{SVMLinear} and \emph{Julia's} \emph{PartitionTree}.

\vskip 6pt

For the FordB dataset (Table~\ref{tab:fordb}), the best performers are: \emph{Random Forest} from \emph{Caret}, \emph{k-NN} from \emph{ScikitLearn}, and \emph{Vote Ensemble} from \emph{Julia}. The worst performers are similar to Tables~\ref{tab:elec}~and~\ref{tab:ref}.

\subsection{Discussion}
Comparing the performances of different classifiers among the four problems indicate that the most difficult problem to classify is the Refrigeration Devices while the easiest one is the Earthquakes problem. As we expected, there is no enough sample for the classifiers to learn the mapping in Refrigeration dataset relative to its feature dimension. One way to alleviate this problem is to remove features that are highly correlated or perform PCA to reduce the feature dimension. This is beyond the scope of the current paper which focuses mainly on the applicability of TSML MLs to different set of problems. 

\vskip 6pt

Among the classifiers, the \emph{Random Forest} from \emph{Caret} is the top performer in all problems. On the other hand, \emph{TreeBag} from \emph{Caret} and \emph{Gradient Boosting} from \emph{ScikitLearn} are the top performers in Refrigeration Devices and Earthquakes, respectively. It is interesting to note that among the \emph{Random Forest} implementations, the top performer is the \emph{Caret's} version written by Breiman\cite{Breiman2001} who was the original author of the said algorithm in \emph{Fortran}. The superior performance of Breiman's \emph{Random Forest} can be highlighted in the FordB classification performance.  \emph{k}NN, the next best classifier is 4\% lower than Breiman's \emph{Random Forest}. In all other cases, the next best performer has almost same performance with  Breiman's \emph{Random Forest}.
 
\vskip 6pt
 
As we expected based on past studies, the different ensemble models dominated the top 5 performers. While not shown in the table, the most consistent worst performer is the \emph{Caret's} \emph{RPart}. This is consistent to Breiman's observation that \emph{Random Forest} performs well by using unstable or inferior ML models in its leaves. In \emph{Caret's} \emph{Random Forest}, the leaves are composed of \emph{RPart} models which has the poorest performance in all problems. The combination of boosting and bagging weak learners such as \emph{RPart} make the \emph{Random Forest} robust from datasets imbalances. 
 
\section{Summary and Conclusion}
Packages for time series analysis are becoming important tools with the rapid proliferation of sensor data brought about by \textbf{IoT}.  We created \textbf{TSML} as a time series machine learning framework which can easily be extended to handle large volume of time series data. 
\textbf{TSML}  exploits the following \emph{Julia} features: multiple dispatch, type inference, custom data types and abstraction, and parallel computations. 

\vskip 6pt

\textbf{TSML} main strength is the adoption of \emph{UNIX} pipeline architecture containing filters and machine learners to perform both preprocessing and modelling tasks.  In addition, \textbf{TSML} uses a common machine learning API for both internal and external ML libraries, distributed and threaded support for modeling, and a growing collection of filters for preprocessing, classification, clustering, and prediction. 

\vskip 6pt

Extending \textbf{TSML} can easily be done by creating a custom data type filter and defining its corresponding \texttt{fit!} and \texttt{transform!} operations which the \textbf{TSML} pipeline iteratively calls for each transformer in the workflow.


\input{bib.tex}

\end{document}

%% file: header.tex

\title{TSML (Time Series Machine Learning)}

\author{Paulito P. Palmes}
\author{Joern Ploennigs}
\author{Niall Brady}
\affil{IBM Dublin Research Lab}

\keywords{Julia, Time Series, Machine Learning, Filters, Feature Extraction, Classification, Prediction, Aggregation, Imputation}

%% file: bib.tex

\bibliographystyle{juliacon}
\bibliography{ref.bib}

%% file: paper.bbl
\begin{thebibliography}{1}

\bibitem{bezanson2017julia}
Jeff Bezanson, Alan Edelman, Stefan Karpinski, and Viral~B Shah.
Julia: A fresh approach to numerical computing.
{\em SIAM review}, 59(1):65--98, 2017.

\bibitem{Breiman2001}
Leo Breiman.
Random forests.
{\em Machine Learning}, 45(1):5--32, Oct 2001.

\bibitem{orchestra2014}
S.~Jenkins.
Orchestra: Heterogeneous ensemble learning for julia.
\url{https://github.com/svs14/Orchestra.jl}, 2014.

\bibitem{combineml2016}
P.~Palmes.
{CombineML}: A package to create heterogeneous ensembles of {ML} from
  {ScikitLearn}, {Caret}, and {Julia}.
\url{https://github.com/ppalmes/CombineML.jl}, 2016.

\bibitem{tsml2019}
P.~Palmes.
{TSML}: Time series machine learning.
\url{https://github.com/IBM/TSML.jl}, 2019.

\bibitem{tsmlextra2019}
P.~Palmes.
{TSMLextra}: External machine learning libs for tsml.
\url{https://github.com/ppalmes/TSMLextra.jl}, 2019.

\bibitem{decisiontree2008}
B~Sadeghi and L.P. Coelho.
Decisiontree: Julia implementation of decision tree and random forest
  algorithms.
\url{https://github.com/bensadeghi/DecisionTree.jl}, 2019.

\end{thebibliography}
